\documentclass{article}
\usepackage{ijcai21}

\RequirePackage[T1]{fontenc} 
\RequirePackage[tt=false, type1=true]{libertine} 
\RequirePackage[varqu]{zi4} 

\RequirePackage[libertine]{newtxmath}
\usepackage[letterspace=-25]{microtype}

\usepackage{natbib}
\usepackage{soul}
\usepackage{url}
\usepackage[hidelinks]{hyperref}
\usepackage[utf8]{inputenc}
\usepackage[small]{caption}
\usepackage{graphicx}
\usepackage{amsmath}
\usepackage{booktabs}
\usepackage{savesym}
\savesymbol{Bbbk}
\savesymbol{openbox}

\usepackage{amssymb}
\usepackage{amsthm}
\usepackage{MnSymbol}
\usepackage{color}
\usepackage{url}
\usepackage{multirow}
\usepackage{xcolor}
\usepackage{dsfont}
\usepackage{algorithm}
\usepackage{algpseudocode}

\restoresymbol{NEW}{Bbbk}
\restoresymbol{NEW}{openbox}

\usepackage[switch]{lineno}

\usepackage{tablefootnote}

\usepackage{subcaption}

\usepackage{todonotes}
\usepackage{xargs}
\usepackage{xcolor}

\newcommandx{\deniz}[2][1=]{\todo[linecolor=blue,backgroundcolor=blue!25,bordercolor=blue,#1]{DG:#2}}
\newcommand{\dg}[1]{\textcolor{orange}{#1}}

\graphicspath{{images}}

\long\def\symbolfootnote[#1]#2{\begingroup%
\def\thefootnote{\fnsymbol{footnote}}\footnote[#1]{#2}\endgroup}
   
\makeatother

\title{
Can Large Language Models perform Relation-based Argument Mining? 
}

\author{Deniz Gorur \and Antonio Rago \and Francesca Toni \\
  Department of Computing, Imperial College London, UK \\
  \texttt{\{d.gorur22,a.rago,ft\}@imperial.ac.uk}
  }

\begin{document}
\maketitle

\begin{abstract}
Argument mining (AM) is the process of automatically extracting arguments, their components and/or relations amongst arguments and components
from 
text. As the number of platforms supporting online debate increases, the need for AM 
becomes ever more urgent, especially in support of 
downstream tasks. 
Relation-based AM (RbAM) is a 
form of AM focusing on identifying 
agreement (support) and disagreement (attack) relations amongst arguments. 
RbAM is a challenging classification task, with existing 
methods 
failing to perform satisfactorily. 
In this paper, we show that general-purpose Large Language Models (LLMs), appropriately primed and prompted, can significantly outperform the best performing (RoBERTa-based) baseline.
Specifically, we experiment with 
two open-source LLMs (Llama-2 and Mistral) 
with ten datasets
. 
\end{abstract}


\section{Introduction}
Argument mining (AM) is the process of automatically extracting 
arguments, their components and/or relations amongst arguments and components
from natural language text~\cite{AMState,AMSurvey}. 
The general AM problem can be split into three main tasks:
1) \emph{argument identification}, involving segmenting text into units and 
determining which 
are argumentative
;
2) \emph{identification of argumentative components}
    , typically involving classifying  claims and/or premises of argumentative text; and 3) \emph{identification of argumentative relations},  
    aiming at determining how different texts are related within argumentative discourse.

As the number of  platforms  supporting  online debate increases, the need for AM becomes ever more urgent~\cite{AMSurvey}. In this paper, we focus on a special form of AM, within the third category, and matching the kind of debate abstractions in platforms such as {\tt kialo.com}, where arguments (textual comments) are connected via {\em support} or {\em attack} argumentative relations. 
Specifically, we will focus on the form of AM 
framed as the following (binary) \emph{relation-based AM} (RbAM) task~\cite{WebContents,IdentifyingArgRelationsUsingDeepLearning,DatasetIndependentRelationExtraction}:\footnote{In \cite{WebContents,IdentifyingArgRelationsUsingDeepLearning}, the task is framed 
as a ternary classification problem, including a third class \emph{no relation}. Here, we focus on the binary version experimented with in \cite{DatasetIndependentRelationExtraction}. }
\emph{
    given a pair $(A,B)$ of 
    texts $A$ and $B$, determine whether 
    $A$ attacks or supports $B$.}
For example, take the three arguments, drawn from the Debatepedia/Procon dataset~\cite{DebatepediaProcon}, $a_1$=`Abortion should be legal', $a_2$=`A baby should not come into the world unwanted', and $a_3$=`Abortion increases the likelihood that women will develop breast cancer'. 
Here, $a_2$ can be deemed to support $a_1$ and $a_3$ to attack $a_1$.

RbAM 
can be used 
to support several downstream tasks, for example, to gather evidence~\cite{WebContents}, 
to determine which online arguments are acceptable~\cite{TweetiesSquabbling}, and 
to analyse divisive issues about new regulations~\cite{AnalyseDivisiveIssues}.
However, 
it is a challenging task, with 
different BERT-based models performing reasonably well on some datasets but individual baselines failing 
to perform well across datasets~\cite{DatasetIndependentRelationExtraction,TransformerBasedModelArgRelation}. 

In this paper, we focus on deploying general-purpose LLMs, with appropriate priming and prompting, to address the RbAM task uniformly across several datasets. In doing so we draw inspiration from recent works showing that LLMs perform significantly better than 
existing baselines on other AM tasks \cite{ExploringLLMinCA,
LLMinLegalArgumentMining,WillItBlend} (see §\ref{sec:related}). 
Overall, our contributions are as follows: 
\begin{itemize}
    \item We provide
    a 
    method for performing RbAM effectively with chat-based LLMs, appropriately, but simply, primed and prompted (see §\ref{sec:method}).
    \item We demonstrate empirically, with a wide-ranging evaluation with ten datasets from the 
    literature (see §\ref{sec:exp}), that our LLM-based method for RbAM outperforms 
    the state-of-the-art RoBERTa baseline for RbAM~\cite{TransformerBasedModelArgRelation} 
     (see §\ref{sec:res}).
\end{itemize}

\section{Related Work}
\label{sec:related}

\paragraph{Relation-Based Argument Mining}
The field of RbAM has received significant 
attention in recent years
~\cite{FiveYearsArgMining}.
%
\citet{ArgRelationClassification} introduced a Joint Inference model
and compared it against baseline methods of logistic regression, attention-based LSTMs, and the EDITS method from \citet{CombiningTextEntailAT}, which recognises textual entailment by calculating the distance between 
arguments.
Their method outperformed the baselines with an $F_1$ score of 65, on 
the Debatepedia/Procon dataset~\cite{DebatepediaProcon}, which we also use (but they do not include the Procon debates).
\citet{IdentifyingArgRelationsUsingDeepLearning} used a deep learning architecture with two separate LSTMs on the embeddings of the 
two arguments in each pair, concatenating the outputs using a softmax layer.
Their method  achieved an $F_1$ score of 89
on the  Web-Content dataset~\cite{WebContents} that we also use.
\citet{DatasetIndependentRelationExtraction} 
used four deep learning architectures with different types of embeddings and compared them against baselines of Random Forests and SVMs.
Their method achieved a best macro $F_1$ score of 54, which performed similarly to the baselines,  
on ten datasets, most of 
which we also use\footnote{We do not use AIFdb 
(\url{https://corpora.aifdb.org/}) as it is not obvious how to map it univocally onto RbAM.}.
Another relevant work is by 
\citet{RelationalFineGrainedArgMining}, 
who experimented 
with 
several variants of LSTMs, CAM-Bert, and TACAM-BERT on the UKP corpus~\cite{UKP} that we also use, 
achieving a best $F_1$ score of 80 with TACAM-BERT. 
Meanwhile, 
\citet{ClassifyingArgRelations} used Logical Mechanisms and Argumentation Schemes,
with, as baselines, TGA Net, Hybrid Net, BERT, BERT+Latent Cross, and BERT+Multi-task Learning.
Their best model achieved an $F_1$ score of 77 with a  dataset also collected from the online debate site Kialo 
as one of our datasets, 
and an $F_1$ score of 80 on 
a similar dataset to Debatepedia/Procon~\cite{DebatepediaProcon}  that we use (but without including the Procon debates).
Finally, \citet{TransformerBasedModelArgRelation} 
evaluated various BERT-based models
against 
LSTMs, achieving 
an $F_1$ score of 70 with RoBERTa-large on the US2016 debate corpus and the Moral Maze multi-domain corpus, both from AIFdb (which we do not use -- see footnote 2).

None of the mentioned approaches to RbAM use LLMs, nor do they achieve the satisfactory performance across datasets 
that we aim for.

\paragraph{Argument Mining 
via Large Language Models}
Recently, the exceptional performance of LLMs across a variety of NLP tasks has led to investigations into their performance in a number of AM tasks. 
\citet{ExploringLLMinCA} tested the capabilities of LLMs  
for 
claim detection, evidence detection, 
stance detection\footnote{This 
deals with classifying the stance of arguments towards topics, whereas RbAM deals with classifying the relation between (two) arguments.}, evidence type classification, and argument generation. They used GPT-3.5-Turbo, Flan-UL2, and Llama 2 13B models for testing, demonstrating that the LLMs perform well in these tasks.
\citet{OptimisingLanguageModelsforArgReasoning} fine-tuned GPT Neo, a pre-trained LLM, to generate, by prompting, natural language arguments 
supporting or attacking a topic argument.
However, work is still to be done before LLMs can be deemed to reason argumentatively, a finding echoed by 
 \citet{Hinton_23}.
Further 
challenges are pointed out by \citet{DetectingFallaciesLLM}, who attempted to use LLMs to detect argumentative fallacies but showed that LLMs did not surpass the performance of the RoBERTa-based Transformer model
.
Meanwhile, \citet{LLMinLegalArgumentMining} focused on the classification of argument components in the legal domain with the GPT-3.5 and GPT-4 models, using a bespoke a few-shot prompting strategy
, showing that the LLMs did not surpass the 
domain-specific 
BERT-based baseline.
More promising results were found in a study of LLMs' potential for generating counter-narratives to counteract online hate speech when supplemented by argumentative strategies and analysis \cite{Furman_23}. Here, the argumentative information, provided by either fine-tuning or priming, was shown to improve the quality of the generated counter-narratives in both English and Spanish.
LLMs' potential for AM was also seen by
\citet{WillItBlend}, 
who used LLMs for argument quality prediction
, amounting to classifying the validity and novelty of a given argument, comprising a premise and a conclusion. 
They achieved best performance using a few-shot learning priming strategy with LLMs for the validity task and a Transformer-based model 
fine-tuned for the novelty task.

Importantly, to the best of our knowledge, 
no study to date considered the use of LLMs for RbAM.

\section{
LLMs for RbAM}
\label{sec:method}

\begin{figure}[ht]
    \centering
    \includegraphics[width=\linewidth]{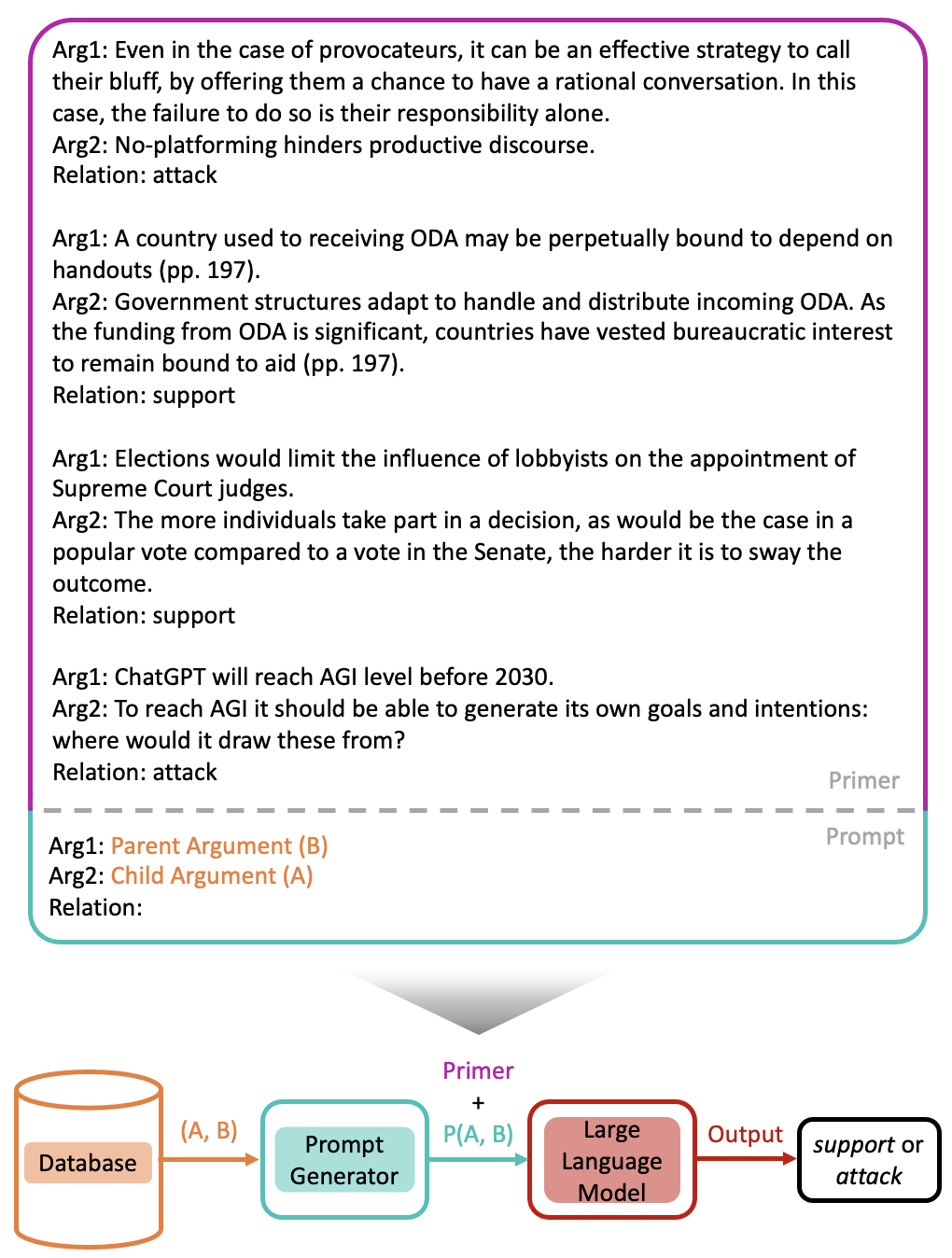}
    \caption{Experimental pipeline with the (few-shot learning) 
    primer and the prompt template P(A,B)
    .}
    \label{fig:pipe}
\end{figure}

Our 
method 
is overviewed in Figure~\ref{fig:pipe}. 
It consists of few-shot priming,
which has shown to perform well with LLMs without the need for fine-tuning \cite{LLMfewshot},  followed by prompting.
The primer uses four labelled examples of attack and support relations between arguments, before we provide an 
example in the prompt for the LLM to classify as attack or support.
The four 
examples 
in the primer are fixed text comprising a parent argument (Arg1), a child argument (Arg2) and the classification of the relation from the child to the parent argument
, as shown in the top, pink 
part of the 
box in Figure~\ref{fig:pipe}.
Then,
the prompt amounts to 
a pair of arguments 
presented as the four in the primer, but without indicating the relation,
as shown in the bottom, turquoise part of the 
box in Figure~\ref{fig:pipe}.
In the experiments, 
the parent and child arguments in the prompt are inputs (from the RbAM datasets described in §\ref{ssec:datasets}). 
Examples of some of these 
prompts are given in Appendix~\ref{app:example_prompts}.



\section{Experimental Set-up}
\label{sec:exp}

We describe the datasets used
, the baseline we compare against 
and the LLMs we experiment with
.\footnote{All our experiments are executed with two RTX 4090 24GB on an Intel(R) Xeon(R) w5-2455X.
In total, it took 112.3 hours to run all the LLM experiments.}

\paragraph{Datasets}
\label{ssec:datasets}
We used ten existing datasets, as follows (see Appendix~\ref{app:dataset_stats} for additional information, including statistics). 
Note that the datasets labelled * directly fit the RbAM task definition (classification of pairs of texts). The dataset labelled $\dagger$ is an extension of a dataset already fitting the RbAM task definition to include additional relations between sentences and topics.
For all these RbAM datasets, we have ignored any relations other than attack and support, given our focus on binary RbAM.  
The other datasets are originally given for different tasks, e.g. to determine relations between sentences and topics  or between  premises and claims: we  adapt them to the RbAM task as discussed in the following.

\emph{Persuasive essays (Essay)}~\cite{PersuasiveEssays} is 
a corpus of annotated 402 persuasive essays
.

\emph{Microtexts* (Mic)}~\cite{Microtexts} 
is a corpus of 112 short texts on controversial issues, with 576 arguments. They were originally written in German and 
then translated to English.

\emph{Nixon-Kennedy debate* (NK)}~\cite{NixonKennedy}  
is a corpus from the 1960 Nixon-Kennedy presidential campaign covering five topics.

\emph{Debatepedia-Procon* (DP)}~\cite{DebatepediaProcon} 
is a corpus extracted from two online debate platforms: Debatepedia 
and Procon
.

\emph{IBM-Debater (IBM)}~\cite{IBMDebater} is 
a dataset containing 55 controversial topics collected from the debate motions database at the International Debate Education Association (IDEA) website
.

\emph{ComArg$\dagger$ 
}~\cite{ComArg} is 
a corpus of user comments collected from Procon 
and IDEA 
where each argument has a stance for or against one of two topics.
For our experiments we 
adapted the dataset so that the parent argument is the topic
. Also, we 
set explicit and vague/implicit attacks to be attacks and vague/implicit and explicit supports to be supports.

\emph{CDCP* 
}~\cite{CDCP} is 
a corpus annotated with only support relations containing 731 user comments on Consumer Debt Collection Practices from the eRulemaking platform
.

\emph{UKP 
}~\cite{UKP} is 
a corpus with arguments obtained from Web documents (including news reports, editorials, blogs, debate forums, and encyclopedias) over eight controversial topics
.
We adapted the parent argument to be `\emph{topic} is good' (e.g. `abortion is good', where abortion is one of the topics).

\interfootnotelinepenalty=10000
\emph{Web-Content* (Web)}~\cite{WebContents}\footnote{To access the dataset, see: \url{https://www.doc.ic.ac.uk/~oc511/ACMToIT2017_dataset.xlsx}} contains 
arguments adapted from the Argument Corpus~\cite{walker-etal-2012-corpus}, plus arguments from news articles, movies, ethics and politics.


\emph{Kialo* 
} 
was collected from the online debate platform Kialo
. Debates (in English) were scraped from Kialo (in 2022) 
on topics related to Politics, Law, and Sports
.

\paragraph{Baseline}
\label{ssec:baselines}
We opted to fine-tune \textbf{RoBERTa}, given its performances in ~\cite{TransformerBasedModelArgRelation}. 
We 
fine-tuned it with 75\% of each dataset separately for 50 epochs (25\% of the datasets were kept for validation), a batch size of 8, and a learning rate of 1e-5.
For each dataset, we selected the best model (over the 50 epochs), 
i.e. that which achieved the highest $F_1$ score on the validation set
.
We then used these candidate 
models (one for each dataset) to perform inference for the other datasets and selected the best  (which turned out to be the one trained on Kialo) as the baseline
(for performances of all these models see Appendix~\ref{app:result_baselines}).

\begin{table*}[htp!]
\centering
\setlength{\tabcolsep}{0.2em}
\begin{tabular}{ccccccc}
\cline{2-7}
 & 
RoBERTa & 
Llama13B & Llama13B-4bit & 
Llama70B-4bit &  
Mistral7B & 
\! Mixtral-8x7B-4bit \! \\ 
\hline
Essays & 
\; 85 / 38 / 80 \; & 
87 / 31 / 82 & 
91 / 36 / 86 & 
94 / 52 / {\bf 90} & 
89 / 42 / 85 &
94 / 43 / 89 \\
Nixon-Kennedy & 
56 / 67 / 62 & 
67 / 12 / 39 & 
66 / 5 / 34 & 
64 / 71 / {\bf 68} &
54 / 68 / 61 & 
66 / 50 / 58 \\
CDCP &
75 / - / 75 & 
87 / - / 87 & 
94 / - / {\bf 94} & 
92 / - / 92 &                        
75 / - / 75 & 
93 / - / 93 \\
UKP &
68 / 81 / 75 & 
70 / 82 / 77 & 
75 / 84 / 80 & 
84 / 89 / {\bf 87}&
78 / 83 / 81 &
81 / 84 / 83 \\
Debatepedia/Procon &
90 / 89 / 90 & 
83 / 71 / 77 & 
84 / 72 / 79 & 
96 / 95 / {\bf 96} &
90 / 89 / 90 &
94 / 93 / 94 \\
IBM-Debater &
85 / 82 / 83 & 
81 / 66 / 75 & 
88 / 82 / 85 & 
94 / 92 / 93 &
89 / 89 / 89 &
95 / 93 / {\bf 94} \\
ComArg &
71 / 74 / 72 & 
68 / 62 / 65 & 
70 / 58 / 65 & 
77 / 56 / 68 &
56 / 71 / 63 &
79 / 73 / {\bf 76} \\
 Microtexts &
73 / 53 / 67 & 
76 / 45 / 67 & 
84 / 41 / 72 & 
81 / 52 / {\bf 73} &
71 / 54 / 67 &
80 / 45 / 70 \\
  Web-Content &
67 / 67 / 67 & 
66 / 63 / 64 & 
68 / 53 / 60 & 
72 / 72 / {\bf 72} &
57 / 72 / 64 &
70 / 66 / 68 \\
Kialo &
- / - / - & 
74 / 56 / 65 & 
75 / 54 / 65 & 
87 / 84 / {\bf 86} &
83 / 83 / 83 &
85 / 82 / 84 \\
\hline
Average &
74 / 61 / 75 & 
76 / 49 / 70 & 
79 / 48 / 72 & 
84 / 66 / {\bf 82} &
74 / 65 / 76 &
84 / 63 / 81 \\
Macro $F_1$ &                       
68 &                               
62 &                                    
64 &                                    
{\bf 75} &                        
70 & 
73 \\ 
\hline
Inference Time (s) & 
0.005 &                               
0.11 &                                    
0.34 &                                    
1.73 &                        
0.06 &
0.28 \\ 
\hline
\end{tabular}
\caption{$F_1$ scores (as a percentage) for  support / attack / both relations in various datasets (rows) for the models used (columns). RoBERTa here is the baseline (
see §\ref{ssec:baselines}) 
and boldface font indicates the best performing model (for both relations) for each dataset. The last row gives the time it takes for a single inference for each model, in seconds.
\label{tab:results}
}
\end{table*}

\paragraph{Large Language Models.}
\label{ssec:llms}
We chose two families of LLMs, both open-source (details are in Appendix~\ref{app:hyper}).
Since LLMs have a huge number of parameters and require a large amount of GPU space, there have been  attempts to reduce the space they take by compressing them to smaller sizes
. For example, GPTQ~\cite{GPTQ} 
uses one-shot weight quantisation based on approximate second-order information to reduce the bit size of each weight in the LLM. 
So, for all three LLMs considered, 
we also experimented with 4bit quantisation (so each weight is stored in 4bits on the GPU) as it had the best trade-off between accuracy and space.

The \textbf{Llama 2} models~\cite{Llama2} 
have been pre-trained with 2 trillion tokens and are generally good at causal language modelling. In our experiments, we used the Llama 2 13B model (and its GPTQ quantised version
) 
which has 13 billion parameters and the Llama 2 70B (GPTQ quantised as the base model needs nearly 140GB of GPU space) which has 70 billion parameters.

The \textbf{Mistral 7B} model
\cite{Mistral} is 
a 7 billion parameter pre-trained and fine-tuned LLM. The model is claimed to perform better than any other open source 13 billion parameter LLM (including Llama 2 13B)
~\cite{Mistral}.
%
The \textbf{Mixtral 8x7B} model
\cite{Mixtral} 
builds on the Mistral 7B model by using 8 of them: for every token, the model selects two of the Mistral 7B models to produce an output and combines them~\cite{Mixtral}. Its performance is claimed to be equal to the Llama 2 70B model~\cite{Mixtral}.
In our experiments, we used the Mistral 7B model and the Mixtral 8x7B model (GPTQ quantised as the base model needs nearly 95GB of GPU space).

\section{Results}
\label{sec:res}

\label{ssec:res_RbAM}


Table~\ref{tab:results} shows the results.\footnote{In the vast majority of cases, the 
LLMs responded with either \emph{attack} or \emph{support}, as expected.
However, 
for 43 instances the LLMs generated 
other labels (see Appendix~\ref{app:extra}), a very small number in comparison with the total number of pairs assessed (159604): we ignored them in the 
results.}
We can see that Llama 70B-4bit 
achieved the highest macro $F_1$ score of 75, outperforming all of the baselines. Also, in seven of the datasets (Essay, NK, UKP, DP, Mic, Web, and Kialo), it achieved 
the highest $F_1$ score of all LLMs (as well as better than all baselines in all of these datasets except two, see Appendix~\ref{app:result_baselines}). 
However, the inference time of 1.73 seconds per argument pair for this model was rather high 
(we believe 
this is not just because it is the biggest model, but also because it is GPTQ quantised).


Mixtral 8x7B-4bit 
performed 
almost as well as Llama 70B-4bit, with a macro $F_1$ score of 73, with  average $F_1$ score for the support labels 
as for Llama 70B-4bit but 
 the average $F_1$ score for the attack labels 3 points lower. 
However, it 
achieved the highest 
$F_1$ scores in two datasets (IBM and ComArg
). Its inference time was (a much lower) 0.28 seconds per argument pair 
(we believe  
it may be faster still if we 
did not use 
quantisation).

Mistral 7B 
performed  well 
given that it is smaller than the other LLMs used, 
achieving a macro $F_1$ score of 70 which 
was better than any of the baselines (see Appendix~\ref{app:result_baselines}). However, it 
did not 
outperform other 
LLMs in any dataset.
Mistral 7B 
was also the fastest, with an inference time of 0.06 seconds per argument pair.

Llama 13B and Llama 13B-4bit  
achieved similar macro $F_1$ scores, 62 and 64, respectively. However, their performance on each dataset 
was varied. Llama 13B-4bit 
performed best on  CDCP, 
which was expected as CDCP only contains support labels and Llama 13B-4bit tends to output support more often. Note that, with GPTQ quantisation, the performance improves. 
They both performed worse than the best baselines (see Appendix~\ref{app:result_baselines}). 
We note that Llama13B-4bit was unexpectedly slower than Llama13B
.

\section{Conclusion and Future Work}
\label{sec:conclusion}
We have introduced a method for the RbAM task using general purpose LLMs, appropriately primed and prompted. 
We showed, with experiments on ten datasets and five open-source LLMs (more than half of which quantised), 
that {\bf Llama 70B-4bit} and {\bf Mixtral 8x7B-4bit} 
surpassed the RoBERTa baseline, with the former outperforming the latter but also bringing the 
downsides of slower inference time and greater GPU 
requirements.

For future work there are many potential avenues, including the following three: 1) We 
could mask the entities in sentences to outline their argumentative structure
, which is shown to improve performance for the argument retrieval task~\cite{ein2020CorpusWideAM}. 2) We plan to work on improving the prediction on the attack relations as LLMs and also baselines performed worse on them. 3) We plan to extend this work for the more challenging (ternary) RbAM task, i.e. determining whether there is a support, an attack or no relation between two arguments.

\section{Limitations}
\label{sec:limitation}
There are some limitations of our work. First, the task that we consider is the (binary) RbAM task (identifying support/attack) whereas, in most real-world applications, it would be a (ternary) RbAM task (identifying support/attack/no relation) as we discussed in §\ref{sec:conclusion}.
Further, the datasets we used are in English: we are not sure if LLMs will perform as well on RbAM in other languages.
GPU limitations affect our selection of small/quantised models, and we were not able to fine-tune any of the LLMs as it was computationally infeasible.

\section{Ethics Statement}
There are potential risks of LLMs such as social bias and generation of misinformation. In this work, we 
only use LLMs to generate a single token which is support/attack, so there are no risks of generating biased or false information.


\bibliographystyle{named}
\bibliography{bibliography, datasets, related}

\newpage

\appendix

\section*{Appendix}

\section{Example Prompts}
\label{app:example_prompts}
\begin{figure}[!htp]
    \centering
    \includegraphics[width=\linewidth]{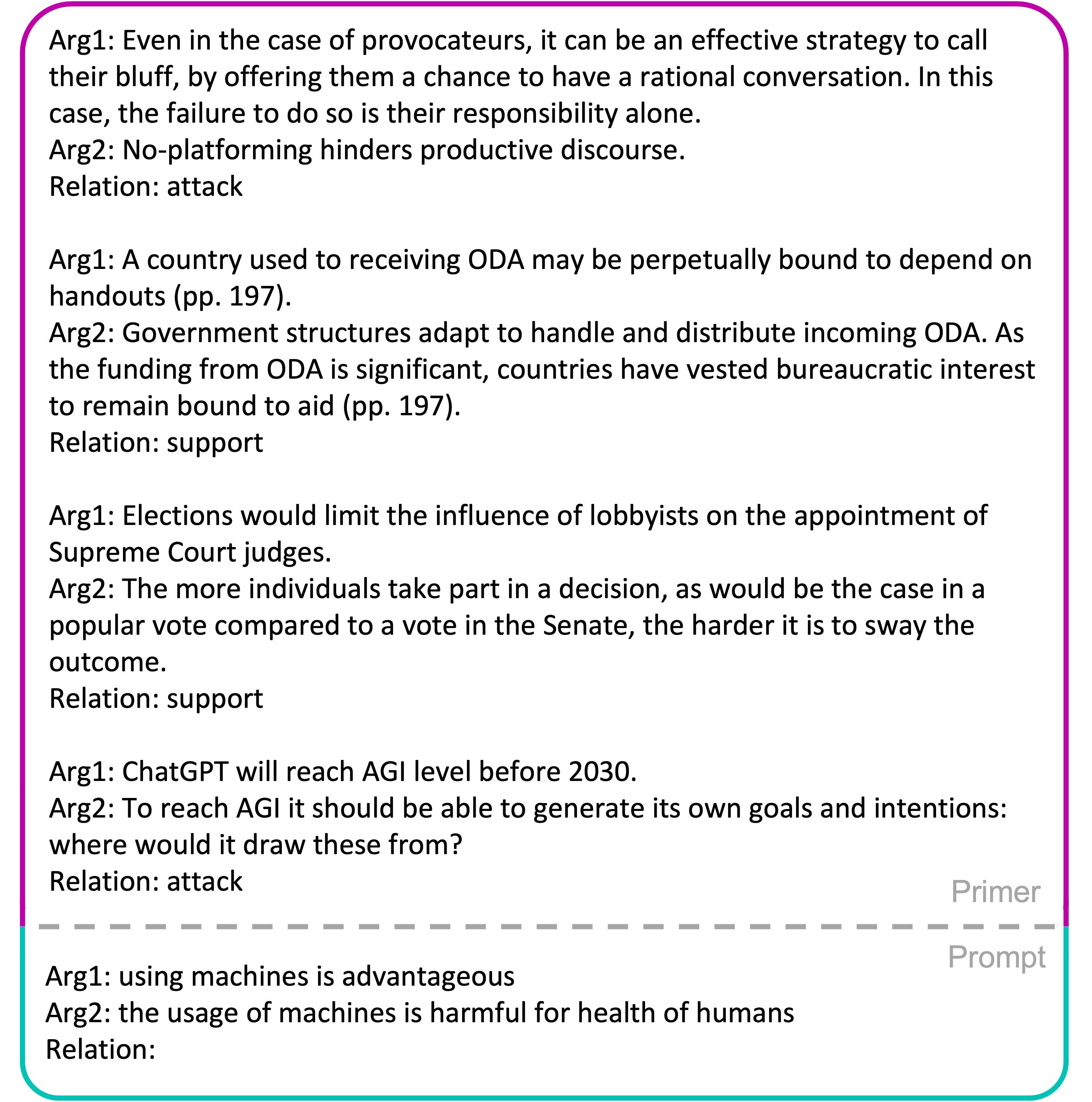}
    \caption{An example prompt drawn from the Essays dataset used in the RbAM experiments.}
    \label{fig:essay-example}
\end{figure}

\begin{figure}[!htp]
    \centering
    \includegraphics[width=\linewidth]{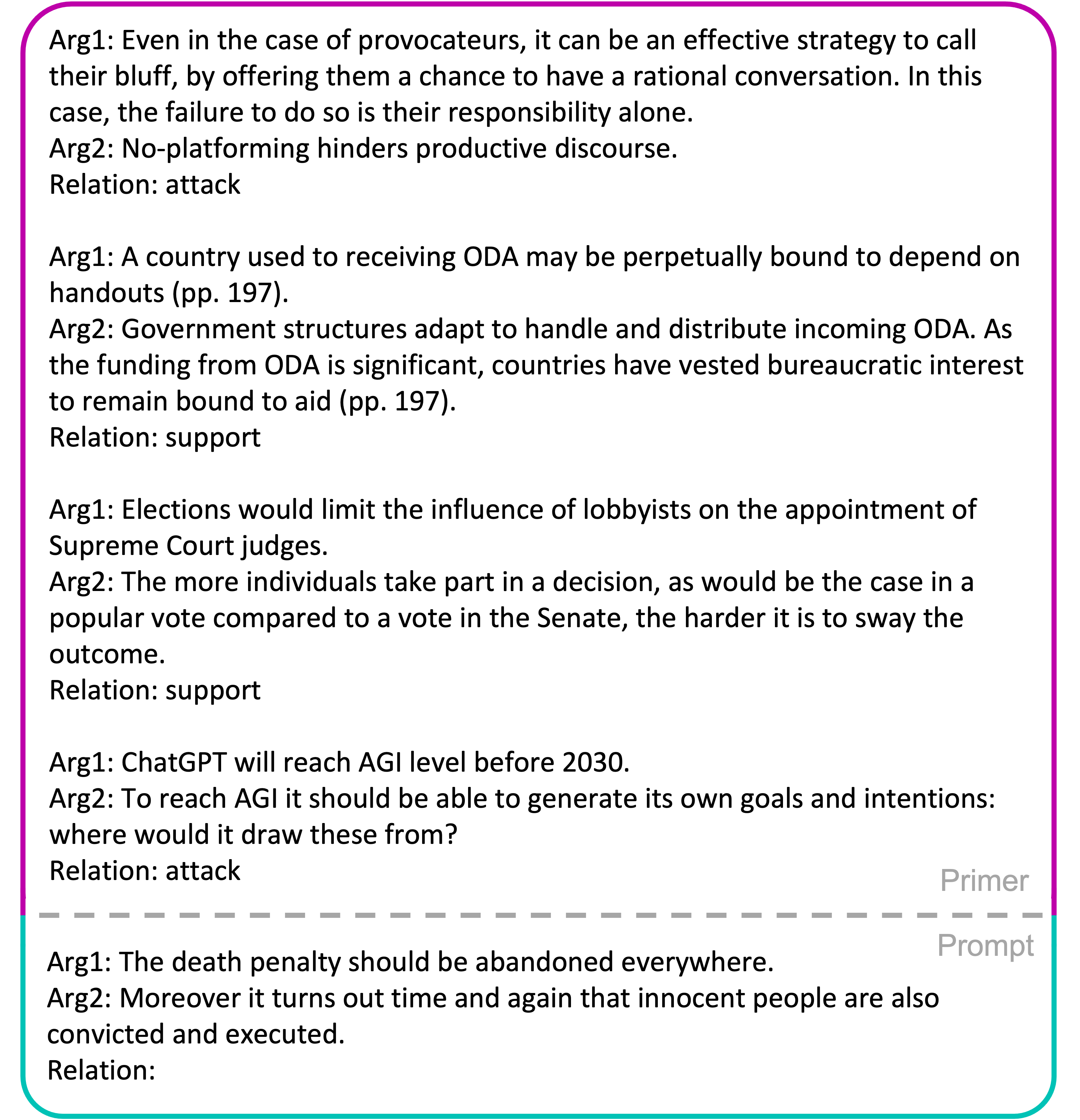}
    \caption{An example prompt drawn from the Microtexts dataset used in the RbAM experiments.}
    \label{fig:mic-example}
\end{figure}

\begin{figure}[!htp]
    \centering
    \includegraphics[width=\linewidth]{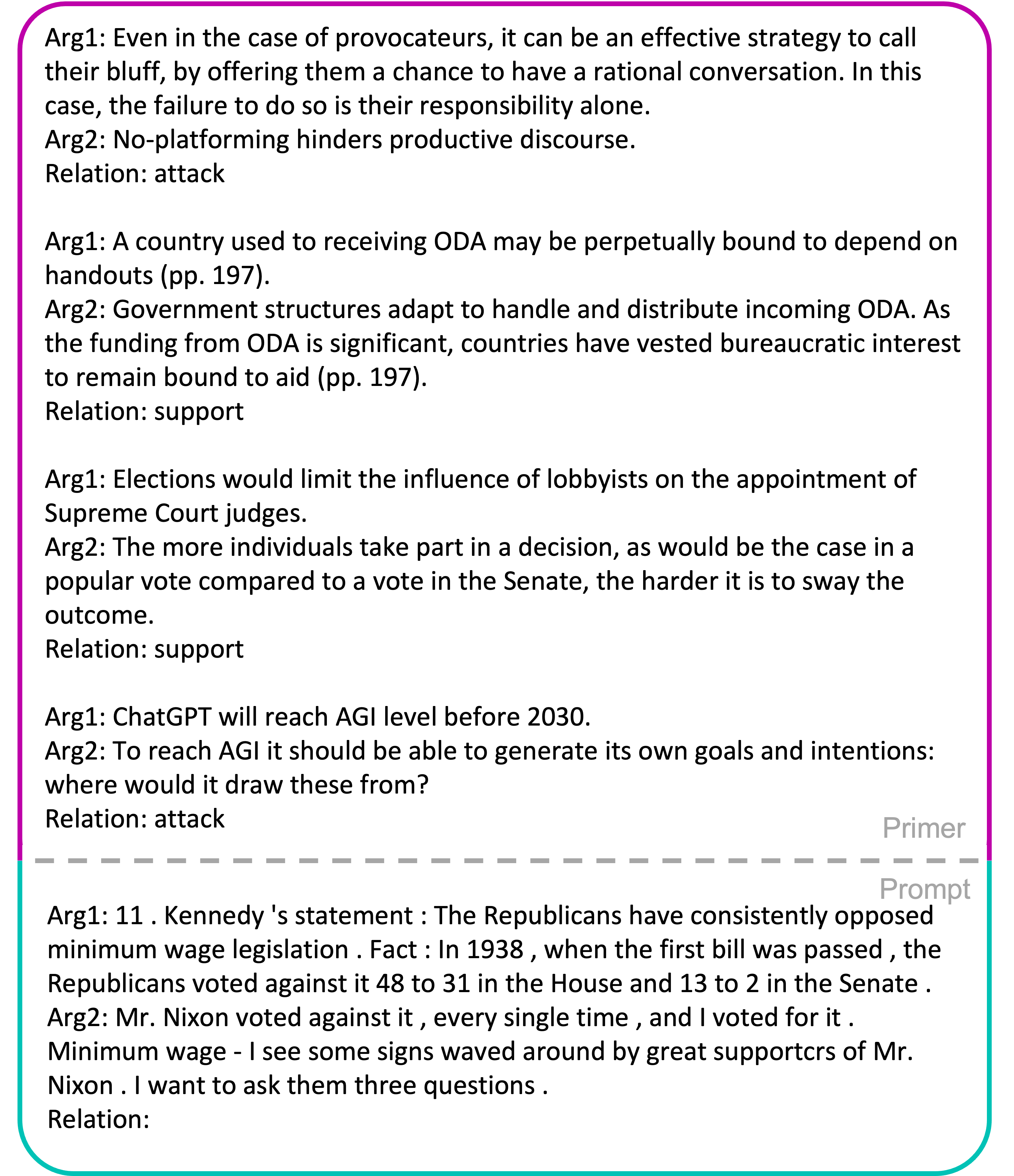}
    \caption{An example prompt drawn from the Nixon-Kennedy dataset used in the RbAM experiments.}
    \label{fig:nk-example}
\end{figure}

\begin{figure}[!htp]
    \centering
    \includegraphics[width=\linewidth]{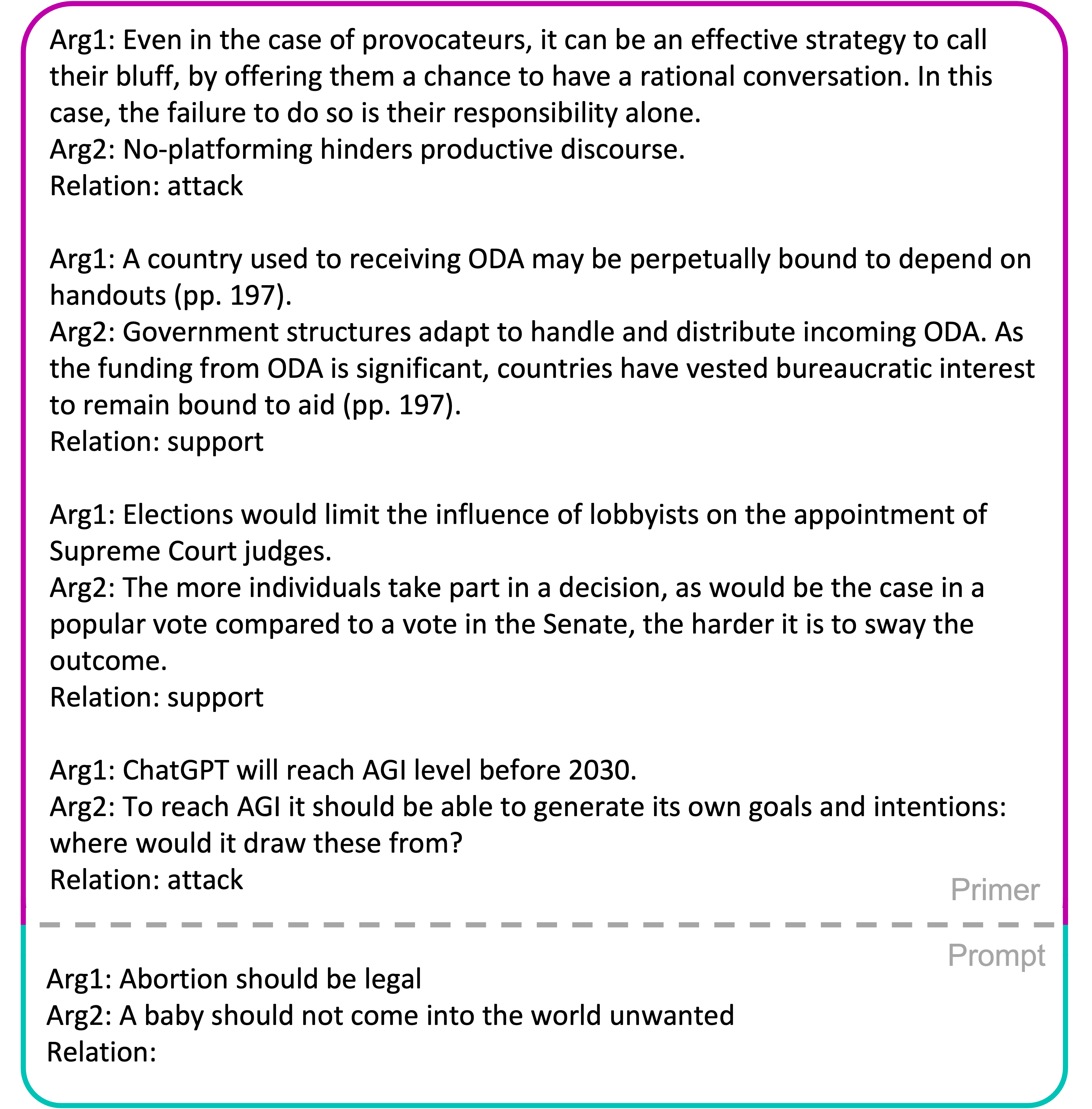}
    \caption{An example prompt drawn from the Debatepedia/Procon dataset used in the RbAM experiments.}
    \label{fig:dp-example}
\end{figure}

\begin{figure}[!htp]
    \centering
    \includegraphics[width=\linewidth]{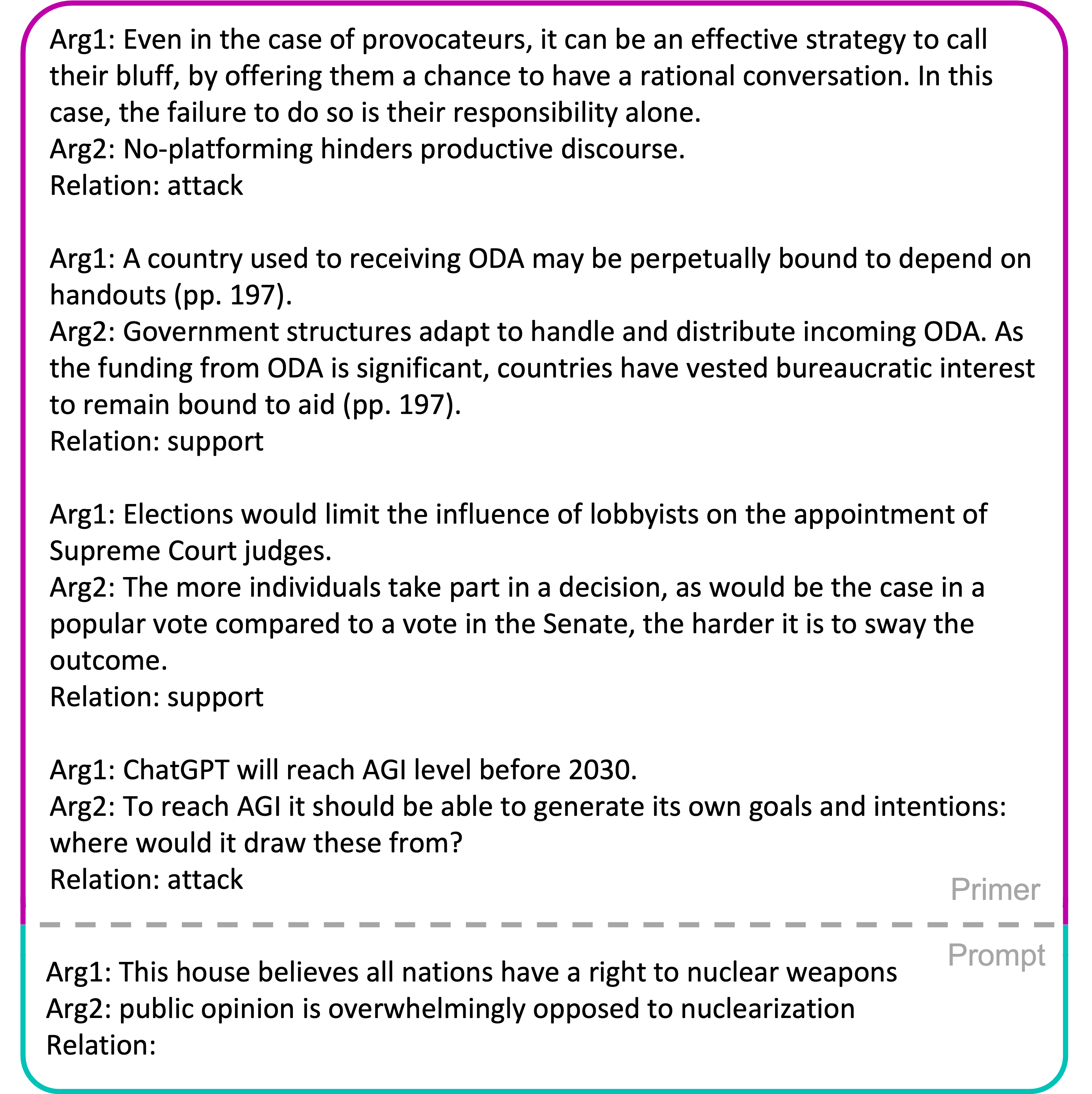}
    \caption{An example prompt drawn from the IBM-Debater dataset used in the RbAM experiments.}
    \label{fig:ibm-example}
\end{figure}

\begin{figure}[!htp]
    \centering
    \includegraphics[width=\linewidth]{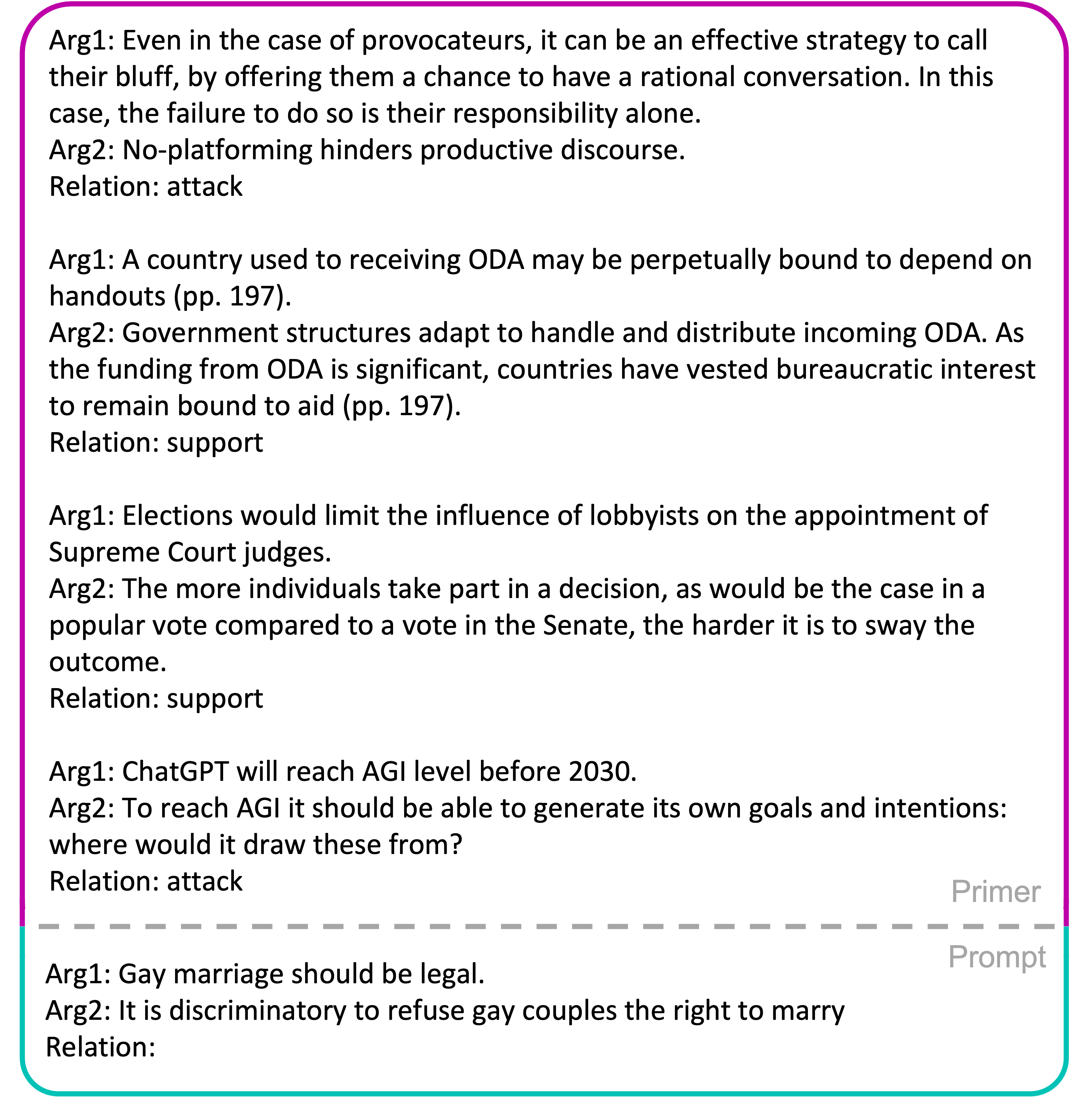}
    \caption{An example prompt drawn from the ComArg dataset used in the RbAM experiments.}
    \label{fig:comarg-example}
\end{figure}

\begin{figure}[!htp]
    \centering
    \includegraphics[width=\linewidth]{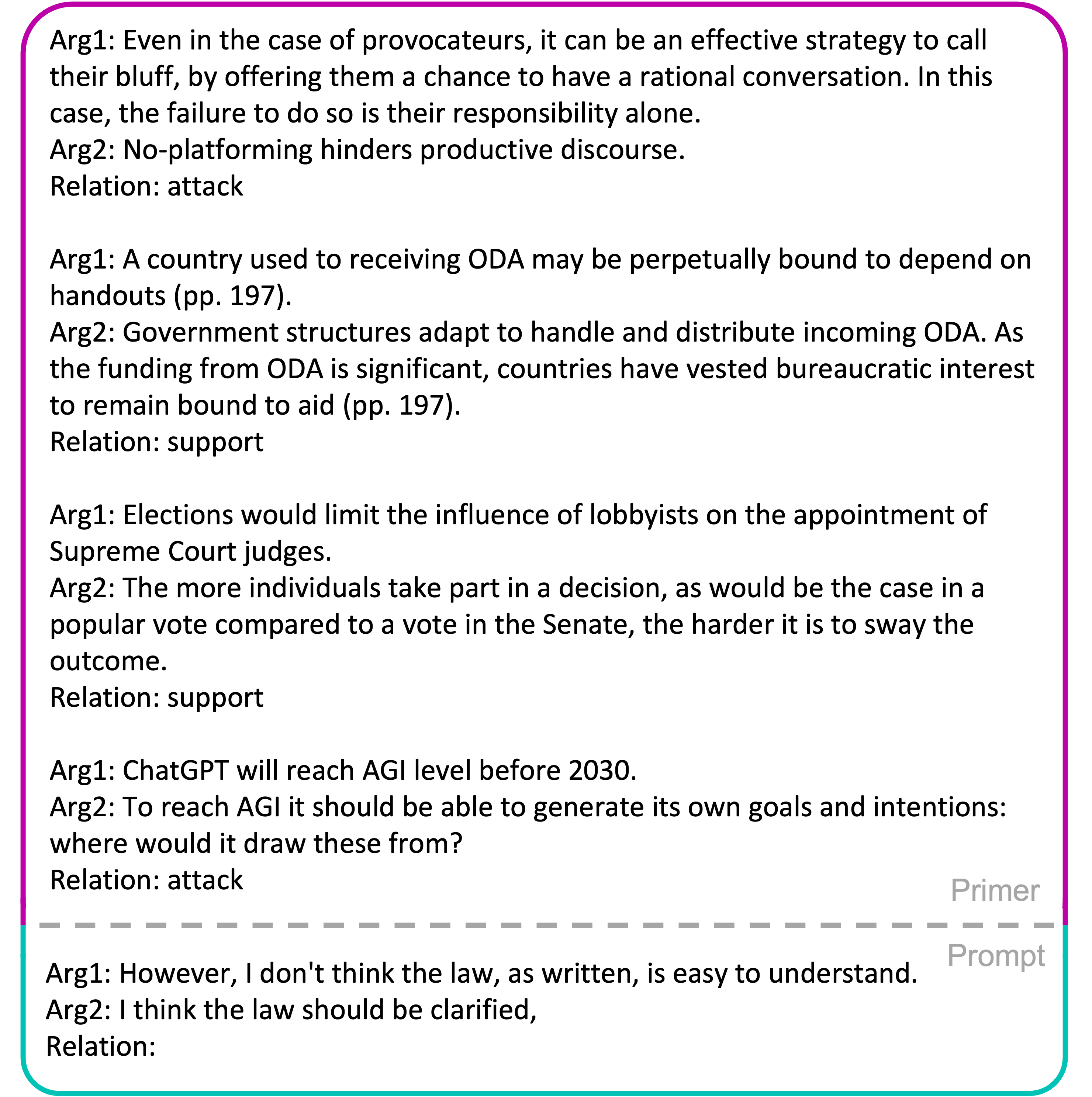}
    \caption{An example prompt drawn from the CDCP dataset used in the RbAM experiments.}
    \label{fig:cdcp-example}
\end{figure}

\begin{figure}[!htp]
    \centering
    \includegraphics[width=\linewidth]{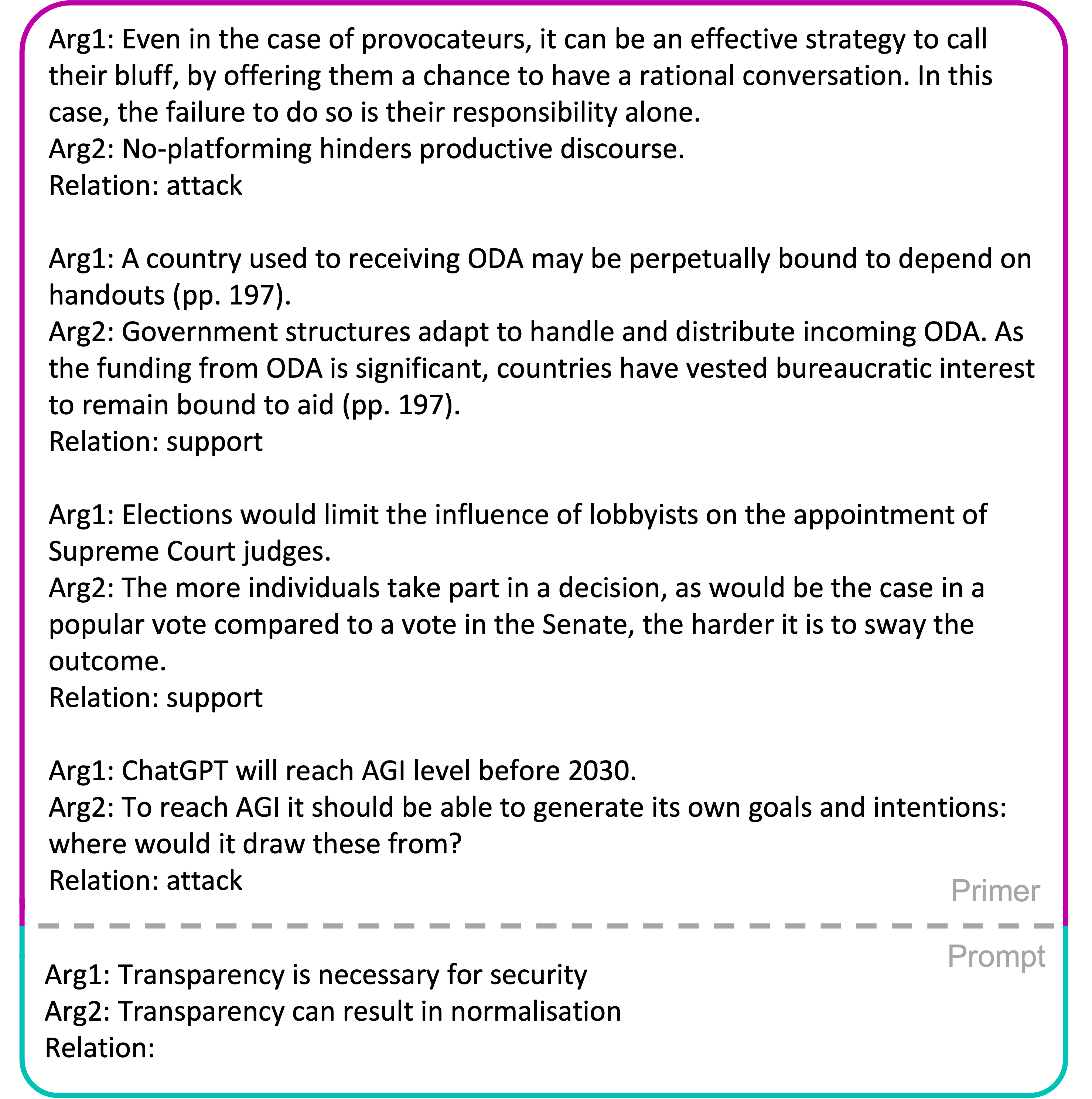}
    \caption{An example prompt drawn from the Web-Content dataset used in the RbAM experiments.}
    \label{fig:web-example}
\end{figure}

In this section we give example prompts generated from each dataset (except the Kialo and UKP datasets as these datasets do not allow us to share them), as seen from Figures~\ref{fig:essay-example},\ref{fig:mic-example},\ref{fig:nk-example},\ref{fig:dp-example},\ref{fig:ibm-example},\ref{fig:comarg-example},\ref{fig:cdcp-example},\ref{fig:web-example}.

\section{Datasets}
\label{app:dataset_stats}
Number of support/attack relations for all these datasets are given in Table~\ref{tab:num_dataset}. This information is important when the $F_1$ scores are calculated. Also, when RoBERTa is fine-tuned on these datasets it is important point how balanced the datasets are.

\begin{table}[!htp]
    \centering
    \setlength{\tabcolsep}{0.4em}
    \begin{tabular}{|c|c|c|c|} \hline  
         Datasets&  \#Support& \#Attack &Total\# \\ \hline  
         Essays&  4841&  497&5338 \\ \hline  
         Microtexts&  322&  121&443 \\ \hline 
         Nixon-Kennedy&  356&  378&734 \\ \hline  
         Debatepedia/Procon&  319&  261&580 \\ \hline  
         IBM-Debater&  1325&  1069&2394 \\ \hline  
         ComArg& 640& 484&1124 \\ \hline   
         CDCP&  1284&  0&1284 \\ \hline  
         UKP&  4944&  6195&11139 \\ \hline  
         Web-content& 1348& 1316& 2664\\ \hline  
         Kialo& 68549& 65355&133904 \\ \hline 
    \end{tabular}
    \caption{Number of support/attack relations in each dataset.}
    \label{tab:num_dataset}
\end{table}

Number of average words and characters for each dataset are given in Table~\ref{tab:stats_dataset}. This kind of statistics help with understanding why all the models under-performed on a specific dataset. For example, in the Nixon-Kennedy dataset the average argument is very long with 103.57 words per argument which contains a lot more information for any model to process and it can be seen that the accuracy is lacking.

\begin{table}[!htp]
    \centering
    \begin{tabular}{|c|c|c|} \hline 
         Datasets&  \multicolumn{1}{c|}{\begin{tabular}[c]{@{}c@{}}Average \#\\ of words\end{tabular}}& \multicolumn{1}{c|}{\begin{tabular}[c]{@{}c@{}}Average \# of\\ characters\end{tabular}}\\ \hline 
         Essays&  14.7 & 87.09\\ \hline 
         Microtexts&  13.58 & 81.3\\ \hline
         Nixon-Kennedy& 103.57 &539.21\\ \hline 
         Debatepedia/Procon&  34.81 & 215.22\\ \hline 
         IBM-Debater&  10.78 & 68.84\\ \hline 
         ComArg&  56.81 & 318.55\\ \hline
         CDCP&  15.4 & 88.11\\ \hline 
         UKP&  15.33 & 83.64\\ \hline 
         Web-content& 19.87 & 112.94\\ \hline 
         Kialo&  21.84 & 135.69\\ \hline
    \end{tabular}
    \caption{Statistical features of each dataset.}
    \label{tab:stats_dataset}
\end{table}

\section{RoBERTa Baselines}
\label{app:result_baselines}
Table~\ref{tab:baseline} shows the results 
for the baselines in the RbAM task, i.e. RoBERTa fine-tuned on each dataset and then evaluated on the remaining datasets.

RoBERTa fine-tuned with the Kialo dataset achieved the highest macro $F_1$ score of 68 and an $F_1$ score better than other baselines in four datasets (NK, UKP, and Web). However, 
note that, since the dataset is large it took 
a long time to fine-tune, specifically 53.73 hours.

RoBERTa fine-tuned with the DP and 
the IBM datasets both achieved a macro $F_1$ score of 66, which came close to the RoBERTa fine-tuned with the Kialo dataset. RoBERTa fine-tuned with the DP dataset achieved a better $F_1$ score than other baselines in three datasets (ComArg, Mic, and Kialo). These datasets are smaller than Kialo and so fine-tuning 
took 0.23 hours for the DP dataset and 0.96 hours for the IBM dataset.

We thus selected RoBERTa fine-tuned with the Kialo dataset as the best baseline, as it performed better than other baselines. 
We note here also that for 
all of the baseline models, a single inference 
took 0.005 seconds for each test sample.

\begin{table*}[htp!]
\centering
\setlength{\tabcolsep}{0.4em}
\hspace*{-1.0cm}
\begin{tabular}{ccccccccccc}
\hline
&
Essay&
NK&
CDCP&
UKP&
DP&
IBM&
ComArg&
Mic&
Web&
Kialo\\ \hline
Essay&
- / - / - &
95 / 5 / 86&
95 / 0 / 86&
71 / 25 / 67&
90 / 42 / 85&
89 / 41 / 84&
94 / 45 / {\bf90}&
79 / 14 / 73&
56 / 16 / 52&
85 / 38 / 80\\ \hline
NK&
65 / 0 / 32&
- / - / -&
65 / 0 / 32&
54 / 46 / 50&
65 / 31 / 47&
60 / 55 / 58&
65 / 4 / 34&
64 / 1 / 32&
46 / 48 / 47&
56 / 67 / {\bf 62}\\ \hline
CDCP&
1 / - / {\bf 1}&
98 / - / 98&
- / - / -&
42 / - / 42&
90 / - / 90&
77 / - / 77&
98 / - / 98&
95 / - / 95&
34 / - / 34&
75 / - / 75\\ \hline
UKP&
67 / 42 / 53&
61 / 28 / 43&
61 / 0 / 27&
- / - / -&
68 / 75 / 72&
73 / 75 / 74&
74 / 67 / 70&
51 / 47 / 49&
58 / 38 / 47&
68 / 81 / {\bf 75}\\ \hline
DP&
75 / 34 / 57&
72 / 23 / 50&
71 / 0 / 39&
62 / 67 / 64&
- / - / -&
84 / 82 / 83&
85 / 78 / 82&
71 / 0 / 39&
61 / 43 / 53&
90 / 89 / {\bf 90}\\ \hline
IBM&
76 / 37 / 59&
72 / 26 / 51&
71 / 0 / 39&
58 / 69 / 63&
82 / 78 / 80&
- / - / -&
87 / 83 / 85&
60 / 33 / 48&
68 / 17 / 45&
85 / 82 / 83\\ \hline
\begin{tabular}[c]{@{}c@{}}Com- \\Arg\end{tabular}&
76 / 36 / 59&
72 / 2 / 42& 
73 / 0 / 41&
59 / 62 / 60&
82 / 73 / {\bf 78}&
73 / 71 / 72&
- / - / -&
72 / 5 / 43&
72 / 3 / 42&
71 / 74 / 72\\ \hline
Mic&
85 / 28 / 70&
83 / 3 / 61&
84 / 0 / 61&
52 / 44 / 50&
83 / 51 / {\bf 74}&
77 / 52 / 71&
83 / 33 / 69&
- / - / -&
60 / 34 / 53&
73 / 53 / 67\\ \hline
Web&
68 / 13 / 41&
67 / 15 / 41&
67 / 0 / 34&
51 / 67 / 59&
65 / 59 / 62&
65 / 60 / 63&
69 / 32 / 51&
61 / 40 / 51&
- / - / -&
67 / 67 / 67\\ \hline
Kialo&              
70 / 18 / 45&
68 / 14 / 42&
68 / 0 / 35&
46 / 63 / 54&
79 / 71 / {\bf 75}&
74 / 73 / 73&
74 / 52 / 63&
67 / 3 / 36&
61 / 36 / 49&
- / - / -\\ \hline
Avg.&
76 / 23 / 57&
76 / 13 / 57&
73 / 0 / 44&
55 / 49 / 57&
78 / 53 / 74&
75 / 57 / 73&
81 / 44 / 71&
69 / 16 / 52&
57 / 26 / 47&
74 / 61 / {\bf 75}
\\ \hline
\begin{tabular}[c]{@{}c@{}}Mac. \\Avg.\end{tabular}&
0.50&
0.45&
0.36&
0.52&
0.66&
0.66&
0.62&
0.42&
0.42&
{\bf 0.68}\\ \hline
\begin{tabular}[c]{@{}c@{}}Train \\Time\\ (in\\ hours)\end{tabular}&
2.14&
0.29&
0.52&
4.47&
0.23&
0.96&
0.45&
0.18&
1.07&
53.73\\ \hline
\end{tabular}
\caption{$F_1$ scores for various datasets (rows) by the RoBERTa baselines, fine-tuned on the datasets (columns), where $F_1$-S stands for the $F_1$ score of the {\em support} relation, $F_1$-A stands for the $F_1$ score of the {\em attack} relation and boldface font indicates the best performing baseline for each dataset. The training time it takes for each RoBERTa model, fine-tuned on the datasets is given in hours in the last row.
}
\label{tab:baseline}
\end{table*}

\section{LLMs}
\label{app:hyper}

The amount of GPU space needed for Llama 13B is 27GB, Llama 13B-4bit is 7.4GB, Llama 70B-4bit is 37GB, Mistral 7B is 15GB, and Mixtral 8x7B-4bit is 25GB.
%
%
For every model, we use the default parameter selection for temperature=0.7, top\_p=1, do\_sample=False. However, max\_new\_tokens=1 as inference time is faster and we only need a single token generated for support/attack. Also, the models that are not quantised are loaded with 16-bit precision for faster inference.

\section{Extra labels}
\label{app:extra}
Across the datasets, there were 43 instances where the LLMs generated additional labels than attack/support.
The additional labels the LLMs generate are different for all the models, as shown in Table~\ref{tab:labels}.

\begin{table*}[htp!]
\centering
\setlength{\tabcolsep}{0.55em}
\begin{tabular}{|l|c|c|c|c|c|}
\hline
& Llama 13B                                                           & Llama 13B-4bit & Llama 70B      & Mistral     & Mixtral                                                                                               \\ \hline
Kialo    & \begin{tabular}[c]{@{}c@{}}compare (25)\\ conflict (1)\end{tabular} & compare (1)    &                & analogy (1) & \begin{tabular}[c]{@{}c@{}}irrelevant (2)\\ contradiction (2)\\ compare(2)\\ contrast(1)\end{tabular} \\ \hline
Essays   &                                                                     &                & paraphrase (1) &             & contradiction (1)                                                                                     \\ \hline
UKP      &                                                                     &                &                &             & contradiction (1)                                                                                     \\ \hline
Web      & reply (1)                                                           &                &                &             &                                                                                                       \\ \hline
ComArg   &                                                                     &                &                &             & paraphrase (1)                                                                                        \\ \hline
CDCP     &                                                                     &                &                &             & paraphrase (1)                                                                                        \\ \hline
NK       & rebuttal (2)                                                        &                &                &             &                                                                                                       \\ \hline
\end{tabular}
\caption{These are the additional labels the LLMs generated (columns) on the datasets (rows). The number in the parentheses represents the number of times the label has been generated.}
\label{tab:labels}
\end{table*}

\end{document}